# Context-based Deep Learning Architecture with Optimal Integration Layer for Image Parsing

Ranju Mandal, Basim Azam, and Brijesh Verma

Centre for Intelligent Systems, School of Engineering and Technology
Central Queensland University, Brisbane, Australia
`{r.mandal,b.azam,b.verma}@cqu.edu.au`

**Abstract.** Deep learning models have been proved to be promising and efficient lately on image parsing tasks. However, deep learning models are not fully capable of incorporating visual and contextual information simultaneously. We propose a new three-layer context-based deep architecture to integrate context explicitly with visual information. The novel idea here is to have a visual layer to learn visual characteristics from binary class-based learners, a contextual layer to learn context, and then an integration layer to learn from both via genetic algorithm-based optimal fusion to produce a final decision. The experimental outcomes when evaluated on benchmark datasets show our approach outperforms existing baseline approaches. Further analysis shows that optimized network weights can improve performance and make stable predictions.

## 1   Introduction

Image parsing is a popular topic in the computer vision research domain with a broad range of applications. Scene understanding, medical image analysis, robotic perception, hazard detection, video surveillance, image compression, augmented reality, autonomous vehicle navigation are some specific applications where image parsing is essential. Because of the performance of deep learning models on benchmark problems in a vast range of computer vision applications, there has been a volume of works focused on designing image parsing models using deep learning. Recent deep CNN-based architectures have the remarkable capability for precise semantic segmentation as well as pixel-wise labeling, however, the context presentations are implicit, and still scene parsing task remains a hard problem.

Two important aspects that make scene parsing challenging: i) clutter objects and shadows in scene images have high intra-class variability ii) presence of low interclass variability due to texture similarity among objects. Recently published architectures [1]-[4] have made progress in pixel-wise labeling using CNN and context information at local and multiple levels. Semantic segmentation processes are mostly enhanced by rich global contextual information [5]-[7]. However, a lack of systematic investigation to utilize local and global contextual information parallelly in single network architecture is visible. Context plays a vital role in complex real-world scene image parsing by acquiring the critical surrounding information, and this captured statistical property can be utilized for predicting class label inference in many applications.

Contextual information can be categorized into two basic types: i) a global context that contains the statistical information derived from a holistic view and ii) local context captures the statistical information of adjacent neighbors. In general, the class label of a pixel in an image should maintain a high uniformity with the local context conveyed by adjacent pixels; match the global context of the entire scene semantics and layout too. Existing approaches are neither able to capture global contextual information from the images nor consider the labels of neighboring object categories. However, deep learning models have the potential to improve performance for semantic segmentation if assimilate such features explicitly. In fact, global and local context representations in deep learning models remain a challenging problem. Few methods have shown misclassification can be reduced by scene context at the class label. Conditional Random Fields (CRFs)-based Graphical model [8], relative location [9], and object co-occurrence statistics-based model [10] have achieved significant performance improvement by the modeling of contextual information in natural scenes.

In this paper, we propose a novel architecture to integrate contextual and visual information in a new way to improve scene parsing. The proposed architecture extracts the visual features and the contextual features to produce precise superpixel-wise class labels and integrates them via a Genetic Algorithm (GA)-based optimization. Both global and local contextual features are represented by computing object occurrence priors of each superpixel. We introduce novel learning methods where multiple weak learners are fused to achieve high performance. The class-specific model classifier considers both; many weak classifiers and boosting algorithms to achieve improved accuracy of class labeling. The final integration layer combines three sets of class-wise probabilities acquired from the classification of superpixels, local context, and global context. A neural network optimized with the GA instead of gradient descent to avoid being trapped into 'local minima' and computes the final class label. The primary contributions of this paper can be summarized as follows:

a) The proposed new robust architecture incorporates both visual features and contextual features and unified them into a single network model. To leverage the binary classification, we use the one vs. all classification technique to obtain a class-wise probability of superpixels. Based on the number of target classes, our one vs. all solution consists of an equal number of binary classifiers for each possible outcome.

b) The model is trained with object co-occurrence and spatial context simultaneously from the training image samples in an unsupervised fashion. The proposed Context Adaptive Voting (CAV) features conserve relative and absolute spatial coordinates of objects by capturing Objects Co-occurrence Priors (OCPs) among spatially divided image blocks.

c) We optimize the weights of the integration layer using a novel fusion method to obtain the final classification label. GA-based optimization is proposed in this layer to overcome the 'local minima' problem.

d) We conduct exhaustive experimentations with benchmark datasets, in particular, Stanford Background Dataset [11] and CamVid [12], and compare the results with state-of-the-art methods.

## 2      Related Works

An up-to-date review of the lately published state-of-the-arts works in image parsing is presented here. We critically reviewed multi-scale and spatial pyramid-based architectures, encoder-decoder networks, models based on visual attention, pixel-labeling CNN, and adversarial settings generative models. Besides, as the GA is incorporated in our integration layer, we explore few recent GA-based optimization techniques on the hyper-parameter solution space to optimize the CNN architectures' weights.

The fundamental attributes of the contemporary methods comprise of CNN-based model and multi-scale context. In contrast to the standard convolution operator, DenseASPP's [1] atrous convolution secure a larger receptive field without compromising spatial resolution and keeps the same number of kernel parameters. It encodes higher-level semantics by acquiring a feature map of the input shape with every output neuron computing a bigger receptive field. RPPNet [2] analyzes preceding successive stereo pairs for parsing result that efficiently learns the dynamic features from historical stereo pairs to ensure the temporal and spatial consistency of features to predict the representations of the next frame. STC-GAN [3] combines a successive frame generation model with a predictive image parsing model. It takes previous frames of a video and learns temporal representations by computing dynamic features and high-level spatial contexts. Zhang et al. [4] proposed perspective adaptive convolutions automatically adapt the sizes and shapes of receptive fields corresponding to objects' perspective deformation that attenuates the issue of inconsistent large and small objects, due to the fixed shaped receptive fields of standard convolutions.

Preliminary approaches obtain pixel-wise class labels using several low-level visual features extracted at pixel-level [13] or a patch around each pixel [14]. Few early network models on image parsing include feature hierarchies and region proposal-based architectures [15], which consider region proposals to produce class label accuracy. However, features that acquire global context achieve better as features at pixel-level fail to capture robust appearance statistics of the local region, and patch-wise features are susceptible to background noise from objects. PSPNet [16] exploits the global context of an image using the spatial pyramid pooling technique to improves performance. A hybrid DL-GP network [17] segments a scene image into two major regions namely lane and background. It integrates a convolutional encoder-decoder network with a hierarchical GP classifier. The network model outperforms SegNet [18] in both visual and quantitative comparisons, however, considers only the pedestrian lane class for evaluation.

RAPNet [20] proposes importance-aware street frame parsing to include the significance of various object classes and the feature selection module extracts salient features for label predictions, and the spatial pyramid module further improves labeling in a residual refinement method. UPSNet [21] tackles the panoptic segmentation task.

A residual network model followed by a semantic segmentation module and a mask R-CNN-based [22] segmentation module, solves subtasks concurrently, and eventually, a panoptic module performs the panoptic segmentation using pixel classification. A deep gated attention network was proposed by Zhang et al. [23] in the context of pixel-wise labeling tasks which captures the scene layout and multi-level features of images. Cheng et al. [24] propose scene parsing using Panoptic-DeepLab architecture based on a fully convolutional method that simultaneously aims the semantic and instance segmentation in a single-shot, bottom-up fashion. RefineNet [26] enables high-resolution prediction by adopting a method based on long-range residual connections. It uses features obtained at multiple levels available along the down-sampling process. The chained residual pooling encapsulates the robust background context, and labeling was further refined by dilated convolutions, multi-scale features, and refining boundaries.

## 3 Proposed Approach

The proposed network architecture is constructed by pipelined stacked layers. As presented in Fig. 1, the first step is involved in computing superpixels level visual features. The class probabilities of superpixels are predicted in the visual feature prediction layer using the visual features and the class-semantic supervised classifiers. The contextual voting layer attains local and global contextual voting features of each superpixel. These contextual voting features are computed using the corresponding most probable class and the Object Co-occurrence Priors (OCPs). Subsequently, all three features are fed into our final integration layer to fuse optimally and produce the final target class. The optimal weights are learned using a small network with a single hidden layer, and GA is incorporated to find the best solution which holds all parameters that help to enhance the classification results. Three main steps involved in the proposed architecture for scene parsing are described below.

### 3.1 Superpixel Feature Computation and Classification

Let $I(v) \in R^3$ be an image defined on a set of pixels $v$, the goal of scene parsing is to set each pixel $v$ into one of the predefined class labels $C=\{c_i| i=1,2,...,M\}$ where $M$ represents the number of attributes. For superpixel-based scene parsing, let $S(v)=\{s_j|j=1,2,...,N\}$ represents the set of superpixels obtained from our segmentation process from an image $I$, and $N$ represents the number of all superpixels, $F^v=\{f_j^v| j=1,2,...N\}$, $F^l=\{f_j^l| j=1,2,...N\}$ represents, and $F^g=\{f_j^g| j=1,2,...N\}$ represent corresponding visual features, local contextual features, and global contextual features respectively, the goal is then set to labeling all image pixels $v$ in $s_j$ into a class $c_i \in C$ and $v \in s_j$, and the conditional probability of giving a correct label $c_i$ for $s_j$ can be formulated as:

$$P(c_i|s_j, W) = P(c_i|f_j^v, f_j^l, f_j^g; W_v, W_l, W_g) \qquad (1)$$

$s.t. \; \Sigma_{1 \leq i \leq M} P(c_i|s_j) = 1$, where, $W=\{W_v, W_l, W_g\}$ denotes weight parameters for all the three features $F_v$, $F_l$, and $F_g$ respectively, and $W$ is learned during the training pro-cess. The final goal is to get a trained model that is able to maximize the sum of conditional probabilities of giving appropriate labels for all superpixels:

$$P(C|S, \theta) = max_{s_j \epsilon S \, \& \, c_i \epsilon C} \, P(c_i|s_j, W) \qquad (2)$$

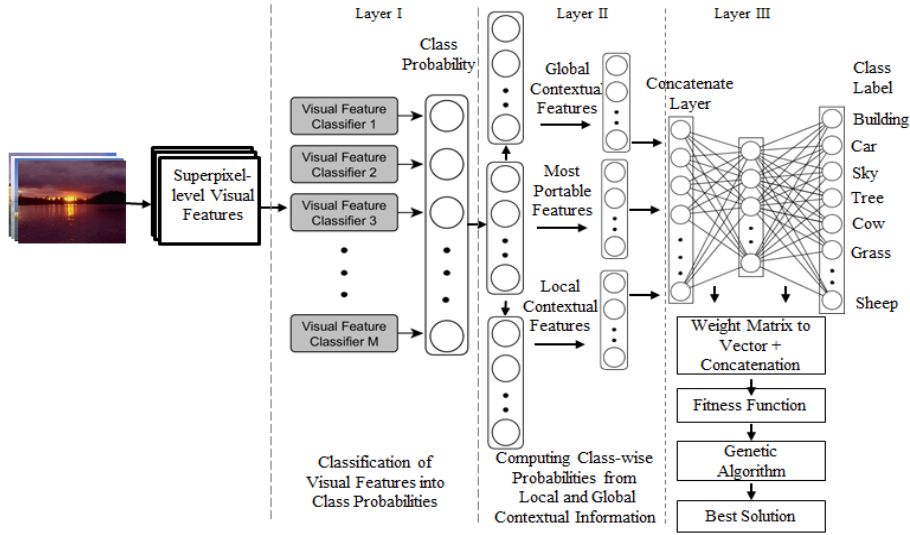

**Fig 1**. The three-layered deep architecture predicts probabilities of every superpixel by class semantic 'One vs. all' classifiers in Layer I. In Layer II, the global and local Context Adaptive Voting (CAV) features are extracted by casting votes from superpixels in spatial blocks and neighboring superpixels respectively using the most probable class and object co-occurrence priors (OCPs). Finally, three class-wise probability vectors obtained from the visual feature, local, and global CAV features are integrated into Layer III to predict the final class label.

### 3.2 Prediction of Visual Features

The visual prediction layer comprises multiple binary (one vs. all) classifiers that take visual features as input to predict the probabilities approximately for all superpixels belonging to each class. It provides a way to leverage binary classification over a multi-class classifier. The resultant probabilities vector serves as an early prediction and becomes a foundation for computing local and global contextual features based on image context. For each superpixel, we obtained a probability vector of size equal to the total class number. A set of binary (one-vs-all) classifiers are developed and trained for an individual class instead of training a single multi-class classifier for all classes. Given a multi-class classification problem with M feasible solutions, a one-vs.-all classification model contains M independent binary classifiers, one binary classifier for every possible outcome. The use of class-specific classifiers provides three benefits: a) class-specific classifier-based architecture mitigates the imbalanced training data problem among classes, especially for an organic dataset containing a batch of sparsely occurring but crucial classes. In public benchmark scene parsing datasets, the distribution of the pixels is presumably heavy-tailed to many common attributes whereas sparse pixels can be available for uncommon classes. In such circumstances, training a multi-class model has the probability of totally disregard uncommon classes and being influenced towards prevalent classes. b) Reinforce class-specific feature selection, which permits the use of the exclusive feature sets for every class. c) Training of multiple class-specific classifiers that proved to be more potent contrarily to train a single multi-class classifier, especially for datasets containing vast class numbers.

### 3.3 Local and Global Context-based Features

The contextual voting layer plays a pivotal role in the proposed architecture as it aims to learn the context-adaptive voting (CAV) features that incorporate both short- and long-range label connections among attributes in images and are robust to the local characteristic of images. The contextual voting layer computes CAV features of each superpixel at local and global context levels. Superpixel-wise predicted class labels and the corresponding object co-occurrence priors (OCPs) are used to compute CAV features. The contextual voting layer first computes global OCPs and local OCPs from the training dataset. Finally uses these two priors with the most probable class to computes global and local CAV features. Subsequently, these three (i.e. global CAV, local CAV, and class probability) vectors are fed to the final integration layer. The main objective of the contextual exploitation layer is to consider the contextual attributes of an object, computing the co-occurrence priors of object classes from the training images, and extract context-adaptive voting features for each superpixel. This layer receives the most probable class information by the initial prediction of visual features from Layer-I and uses it with OCPs to obtain CAV features. The local and global context priors that are learned are reflected in the contextual features, and we obtain these features by casting votes for image superpixels during testing as presented in Equations (3-5).

$$V^l(C|s_j) = \psi^l(P^v(C|s_j), OCP) \tag{3}$$

$$V^g(C|s_j) = \psi^g(P^v(C|s_j), OCP) \tag{4}$$

$$V^{con}(C|s_j) = \zeta\left(\overbrace{(V^l(C|s_j)}^{local}, \overbrace{(V^g(C|s_j)}^{global}\right) \tag{5}$$

where, $\psi^l$ and $\psi^g$ indicate the voting functions for local and global context, respectively, and $\zeta$ is the function that fuses them. The contextual features adaptively obtain the superpixels level dependencies within an image.

It was observed that both visual and contextual features reveal crucial information in the scene parsing problem. Contextual features are regularized to class probabilities to integrate with the class probabilities obtained from the visual classifier. The unification layer models the correlation of class probabilities obtained from visual and contextual features.

$$P(C|s_j) = \mathcal{H}\left(\overbrace{(P^{vis}(C|s_j),}^{visual\ features}\ \overbrace{(P^{con}(C|s_j)}^{contextual\ features}\right) \tag{6}$$

$$P^T(C|s_j) = U_{1 \leq i \leq M} P(c_i|f_i^T; W^T) \tag{7}$$

where $\mathcal{H}$ indicates the joint modeling function for both class probabilities $P^T(C|s_j)$ of $s_j$ using features $f_i^T$ and the weights $W^T$, $T \epsilon \{vis, con\}$.

1) Visual features extracted from superpixels in the visual feature layer are utilized to train several class-specific classifiers to get a superpixel-wise approximate probability of belonging to each class. For the $j^{th}$ superpixel $s_j$, and its class probability for $i^{th}$ class is $c_j$:

$$P^{vis}(c_j|s_j) = \phi_i(f_{i,j}^{vis}) = fn(w_{1,i}f_{i,j}^{vis} + b_{1,i}) \tag{8}$$

where $f_{i,j}^{vis}$ represents visual features of $s_j$ obtained for the i[th] class $c_i$, $\phi_i$ is trained binary classifier at the initial stage for $c_i$, $fn$ represents prediction function of $\phi_i$, trainable weights, constants parameters are $w_{1,i}$ and $b_{1,i}$ respectively. For all classes $M$, the probability vector we obtained for $s_j$:

$$P^{vis}(C|s_j) = [P^{vis}(c_1|s_j), \dots, P^{vis}(c_i|s_j), \dots, P^{vis}(c_M|s_j)] \tag{9}$$

The $P$ vector holds the likelihood of superpixel $s_j$ belonging to all classes C, and $s_j$ is assigned to the class that holds max probability:

$$s_j \epsilon \hat{c} \quad if\ P^{vis}(\hat{c}|s_j) = \max_{1 \leq i \leq M}(P^{vis}(c_i|s_j) \tag{10}$$

2) The votes for the contextual label of all classes for $s_j$ are computed using the contextual features:

$$V^{con}(C|s_j) = [V^{con}(c_1|s_j), \dots, V^{con}(c_i|s_j), \dots, V^{con}(c_M|s_j)] \tag{11}$$

### 3.4   Integration Layer

The main objective of our integration layer is to determine a class label for each superpixel through learning optimal weights. The optimal weights seamlessly integrate class probabilities derived from the most probable class, local CAV features, and global CAV features. The final layer facilitates learning a set of class-specific weights to best represent the correlations among three predicted class probabilities and the class label. These weights are learned to inherently account for contributions of visual and contextual features to determine superpixel-wise class labels for testing images. The integration layer in the proposed approach has one hidden layer. The GA-based method [19, 27, 28] is applied on hyper-parameter solution space to optimize the network architectures' weights. As gradient descent-based weight optimization of Artificial Neural Networks is susceptible to multiple local minima, here we applied a standard GA to train the weights to overcome the 'local minima' problem. In GA, a single solution contains the full network weights of the proposed integrated layer, in the final step of our system architecture, and the steps are repeated for many generations to obtain the optimal weights. GA process is shown below in Algorithm 1.

---
**Algorithm 1.** GA Process for Network Optimization

---
**Input:** Class probability vectors: visual ($F_v$) local ($F_l$) and global ($F_g$), generations $T$, population N, mutation and crossover probabilities $p_M$ and $p_C$, mutation parameter $q_M$, and crossover parameter $q_C$.
**Initialization:** generate a set of randomized individuals and recognition accuracies
**Output:** Class labels

---
**for** $t = 1, 2, ..., T$ **do**
    **Selection:** produce a new generation using the Russian roulette process
    **Crossover:** for each pair perform crossover with probability $p_C$ and parameter $q_C$;
    **Mutation:** do mutation with probability $p_M$ and parameter $q_M$ for every non-crossover individual
    **Evaluation:** compute the recognition accuracy for each new individual
**end for**

---
**Output:** $N$ sets of individuals in the final optimal weight matrix $W = \{W_1, W_2, ..., W_N\}$ with their recognition accuracies, where $N$ presents the number of attributes or classes.

---

The iteration of our genetic optimization iteration starts from this original population, and a fitness evaluation is performed at first. The Fitness evaluation checks where the models are on the optimization surface and determining which of the models perform best. The proposed network learns an optimized set of weights for all its layers. Indeed, the estimation of a globally optimized weight set is often very challenging for a deep architecture with each layer comprising multiple sub-layers. The model parameters are trained layer by layer. The proposed network learns weights in each layer independently to generate an approximation of the best weights.

## 4    Results and Discussion

A detailed comparative study with state-of-the-art scene parsing models is presented in this section, and here we draw a comparison from a performance perspective on two popular benchmark datasets.

### 4.1    Dataset

To evaluate the proposed approach, we adopt the widely used Stanford Background Dataset (SBD) [11] and the CamVid dataset [12]. The SBD contains 715 outdoor scene images and 8 object classes. Each image contains at least one foreground object and 320 × 240 pixels dimension. 572 images are randomly selected for training and 143 images for testing in each fold. The CamVid dataset [12] contains pixel-level ground truth labels. The original resolution of images was downsampled to 480 × 360, and semantic classes were reduced to 11 for our experiments to follow previous works.

### 4.2    Training Details

The proposed architecture is implemented in a MATLAB environment. Experiments are conducted on an HPC cluster, and we assigned a limited resource of 8 CPUs and 64 GB of memory for the experiment. The initial learning rate is 10-4, and 0.1 is the exponential rate of decays after 30 epochs. We initialized our initial population with random initialization and the number of resultant solutions to 8. The number of solutions to be selected as parents in the mating pool was set to 4, the number of generations was set to 1000, and the percentage of genes to mutate was 10. We considered single-point crossover operator, random mutation operator, and kept all parents in the next population. We observed our model convergence up to 1000 generations to find out the optimal solutions.

### 4.3 Evaluation

Table I shows the accuracies of the proposed architecture along with the accuracies reported by previous methods. The proposed network achieved global accuracies of 86.2% and a class accuracy of 85.5%. Table II shows the class accuracy we obtained on the SBD [11]. We obtained 73% and 73.6% mIOU (see. Table III, mIOU is calculated to compare with previous approaches) using MLP and SVM respectively as a classifier at the first layer on the CamVid dataset. The qualitative results obtained on the SBD are presented in Fig. 3. The results demonstrate that our approach successfully predicts object pixels with high precision.

**Table 1.** Performance (%) comparisons with previous state-of-the-art approaches on **Stanford Background Dataset**

| Method | Pixel Acc. | Class Acc. |
|---|---|---|
| Gould et al. [11] (2009) | 76.4 | NA |
| Kumar et al. [30] (2010) | 79.4 | NA |
| Lempitsky et al. [31] (2011) | 81.9 | 72.4 |
| Farabet et al. [8] (2013) | 81.4 | 76.0 |
| Sharma et al. [32] (2015) | 82.3 | 79.1 |
| Chen et al. [33] (2018) | 87.0 | 75.9 |
| Zhe et al. [29] (2019) | 87.7 | 79.0 |
| Proposed Approach | 86.2 | 85.5 |

### 4.4 Comparative Study with Previous Approaches

Table II shows that our approach achieved 86.2% accuracy at par with the recent methods on the SBD, and the class accuracy of 85.5% outperformed previous accuracies (79%). On the CamVid dataset, we obtained 73.6% mIoU that is comparable to the contemporary accuracies, and the comparison is presented in Table IV.

**Table 2.** Class wise performance (%) comparisons with previous approaches on **Stanford Background Dataset**

| Method | Global | Avg. | Sky | Tree | Road | Grass | Water | Bldg. | Mt. | Frgd. |
|---|---|---|---|---|---|---|---|---|---|---|
| Gould et al. [11] | 76.4 | 65.5 | 92.6 | 61.4 | 89.6 | 82.4 | 47.9 | 82.4 | 13.8 | 53.7 |
| Munoz et al. [34] | 76.9 | 66.2 | 91.6 | 66.3 | 86.7 | 83.0 | 59.8 | 78.4 | 5.0 | 63.5 |
| Ladicky et al. [35] | 80.9 | 70.4 | 94.8 | 71.6 | 90.6 | 88.0 | 73.5 | 82.2 | 10.2 | 59.9 |
| **Proposed Method** | **86.2** | **85.0** | **95.4** | **80.4** | **91.6** | **85.1** | **80.2** | **86.8** | **86.7** | **73.9** |

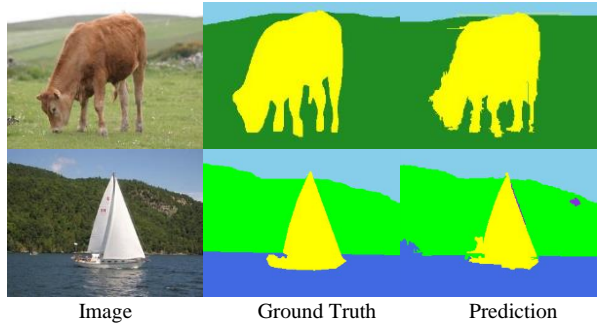

Image            Ground Truth           Prediction

**Fig. 2.** Qualitative results obtained on Stanford Background Dataset (best view in color mode). Original, ground truth, and predicted result images are presented column-wise.

**Table 3.** Performance (%) comparisons with recently proposed approaches on the **CamVid** dataset

| Method | Pretrained | Encoder | mIoU (%) |
| --- | --- | --- | --- |
| SegNet [18] | ImageNet | VGG16 | 60.1 |
| Dilate8 [25] | ImageNet | Dilate | 65.3 |
| BiSeNet [7] | ImageNet | ResNet18 | 68.7 |
| PSPNet [16] | ImageNet | ResNet50 | 69.1 |
| DenseDecoder [6] | ImageNet | ResNeXt101 | 70.9 |
| VideoGCRF [11] | Citscapes | ResNet101 | 75.2 |
| Proposed Approach (MLP) | NA | NA | 73.0 |
| Proposed Approach (SVM) | NA | NA | 73.6 |

## 5   Conclusion and Future Works

The proposed end-to-end deep learning architecture computes object co-occurrence priors from the training phase and taking advantage of the CAV features. It captures both short- and long-range label correlations of object entities in the whole image while capable to acclimate to the local context. The GA optimization finds the optimized set of parameters for the integration layer. We demonstrate that overall performance is comparable with the recent methods. The integrated layer optimization will also be further investigated by incorporating multi-objective optimization.

## 6      Acknowledgements

This research was supported under Australian Research Council's Discovery Projects funding scheme (project number DP200102252).